%% file: main.tex
\documentclass[conference]{IEEEtran}
\IEEEoverridecommandlockouts
\usepackage{cite}
\usepackage{amsmath,amssymb,amsfonts}
\usepackage{algorithmic}
\usepackage{graphicx}
\usepackage{textcomp}
\usepackage{xcolor}
\def\BibTeX{{\rm B\kern-.05em{\sc i\kern-.025em b}\kern-.08em    T\kern-.1667em\lower.7ex\hbox{E}\kern-.125emX}}

\newcommand{\bfsection}[1]{\vspace*{0.1cm}\noindent\textbf{#1.}}

\begin{document}

\title{The XPRESS Challenge: \\ Xray Projectomic Reconstruction \\ -- Extracting Segmentation with Skeletons}

\author{
  \hspace{-1.3cm}
  \begin{tabular}[t]{c}
    Tri Nguyen$^1$, Mukul Narwani$^1$, Mark Larson$^2$,  Yicong Li$^3$, Shuhan Xie$^2$, Hanspeter Pfister$^3$, Donglai Wei$^4$, \\ Nir Shavit$^5$, Lu Mi$^5$, Alexandra Pacureanu$^{6}*$, Wei-Chung Lee$^{7}*$, Aaron T. Kuan$^{1}*$\\
    \\
    $^1$Department of Neurobiology, Harvard Medical School\\ 
    $^2$College of Science, Northeastern University\\ 
    $^3$John A. Paulson School of Engineering and Applied Sciences, Harvard University\\
$^4$Department of Computer Science, Boston College\\
$^5$Computer Science \& Artificial Intelligence Laboratory, MIT\\
$^6$ESRF, The European Synchrotron\\
$^7$F.M. Kirby Neurobiology Center, Boston Children's Hospital, Harvard Medical School
    \\
\end{tabular}
}

\maketitle



\let\thefootnote\relax\footnotetext{\leftline{* Corresponding to}}
\let\thefootnote\relax\footnotetext{\leftline{\texttt{joitapac@esrf.fr, wei-chung\_lee@hms.harvard.edu,}}}
\let\thefootnote\relax\footnotetext{\leftline{\texttt{Aaron\_Kuan@hms.harvard.edu}}}

\input{1-abstract}


\input{2-introduction}

\input{3-data-challenge}

\input{4-baseline-evaluation}

\input{5-plan-schedule}

\newpage
\input{7-reference}

\input{6-biosketch}

\end{document}

%% file: 1-abstract.tex
\begin{abstract}

The wiring and connectivity of neurons form a critical structural basis for the function of the nervous system. Advances in volume electron microscopy (EM) and image segmentation have enabled mapping of circuit diagrams (connectomics) within local regions of the mouse brain. However, applying volume EM over the whole brain is not currently feasible due to technological challenges. As a result, comprehensive maps of long-range connections between brain regions are lacking. 

Recently, we demonstrated that X-ray holographic nano-tomography (XNH) can provide high-resolution images of brain tissue at a much larger scale than EM. In particular, XNH is well-suited to resolve large, myelinated axon tracts (white matter) that make up the bulk of long-range connections (projections) and are critical for inter-region communication. Thus, XNH provides an imaging solution for brain-wide projectomics. However, because XNH data is typically collected at lower resolutions and larger fields-of-view than EM, accurate segmentation of XNH images remains an important challenge that we present here.

In this task, we provide volumetric XNH images of cortical white matter axons from the mouse brain along with ground truth annotations for axon trajectories. Manual voxel-wise annotation of ground truth is a time-consuming bottleneck for training segmentation networks. On the other hand, skeleton-based ground truth is much faster to annotate, and sufficient to determine connectivity. Therefore, we encourage participants to develop methods to leverage skeleton-based training. To this end, we provide two types of ground-truth annotations: a small volume of voxel-wise annotations and a larger volume with skeleton-based annotations. The participants will have the flexibility to use either or both of the provided annotations to train their models, and are challenged to submit an accurate voxel-wise prediction on the test volume. Entries will be evaluated on how accurately the submitted segmentations agree with the ground-truth skeleton annotations. 
\end{abstract}

%% file: 2-introduction.tex
\section{Introduction}
\label{sec:intro}

\begin{figure*}[t]
\begin{center}

\includegraphics[width=0.95\linewidth]{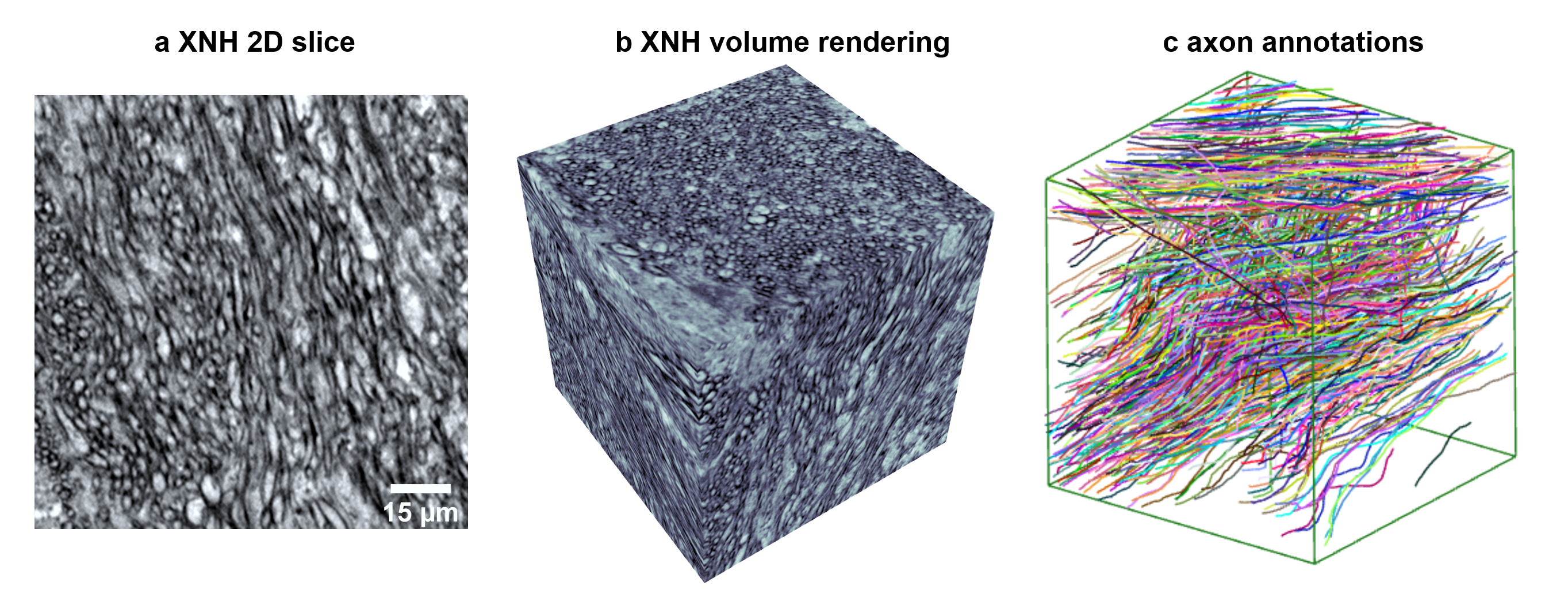}
\end{center}
\vspace{-10pt}
\caption{\textbf{(a)} Example X-ray Nano-Holography (XNH) image from cortical white matter. Data was collected with 100 nm voxel size and corresponds to approximately 200 nm real resolution. Individual myelinated axons, which comprise long-distance connections between brain regions, are clearly resolved. The field-of-view shown is 23.3 µm across. \textbf{(b)} Volume rendering of 3D XNH dataset. XNH data is reconstructed via tomography and has true isotropic resolution. With XNH it is possible to rapidly image large volumes of tissue without thin-sectioning. The field-of-view shown is 23.3 µm across. \textbf{(c)} Axon trajectory annotations provided for the XPRESS challenge. We manually annotated more than 1800 axons within the XNH datasets to reconstruct axon trajectories. The goal of the XPRESS challenge is to automatically segment these axon trajectories with high accuracy.}
\label{fig:dataset}
\end{figure*}

Recent technical advances in 3D electron microscopy (EM)\cite{phelps_reconstruction_2021,eberle_high-resolution_2015, yin_petascale_2020, xu_enhanced_2017} and automated image segmentation \cite{naito2017identification,zaimi2018axondeepseg,mesbah2016deep,pollack2022axon,motta2019dense,plebani2022high} have made it possible to map the complete nervous system of the fruit fly \textit{D. melanogaster} at the level of individual synaptic connections \cite{phelps_reconstruction_2021,zheng_complete_2018,ohyama_multilevel_2015,scheffer_connectome_2020}. However, brute-force scaling of these methods to the size of the mouse brain remains prohibitively costly in time and resources. As a result, only isolated micro-circuits have been mapped in mammalian brains\cite{Kuan2022.04.13.488176,2021.07.28.454025,lee_anatomy_2016,motta2019dense,kasthuri_saturated_2015}.
It has been recently demonstrated that synchrotron-based X-ray nano-holography (XNH) can resolve neuronal wiring at larger scales than EM \cite{kuan2020,bosch_functional_2022}, albeit with slightly lower resolution. XNH is particularly well-suited to map long-range connections, which are composed of thick, myelinated axons in the white-matter regions of the brain (Fig. \ref{fig:dataset}a). White-matter connectivity is critically important for coordination of different brain circuits, with implications for cognition and psychiatric diseases \cite{stafford_immunosignature_2014,whitesell_regional_2021}. With XNH imaging, it is now feasible to comprehensively map long-range connectivity in the mouse brain with single-axon resolution without serial EM sectioning. 

Generating such a “projectome” will require automated axon segmentation algorithms. Several traditional image processing methods have been proposed, such as thresholding and morphological operations \cite{cuisenaire1999automatic}, watershed \cite{wang2012segmentation}, region growing \cite{zhao2010automatic}, and active contours \cite{begin2014automated,zaimi2016axonseg}. More recently, deep learning based techniques have been leveraged to tackle the axon segmentation problem \cite{naito2017identification,zaimi2018axondeepseg,mesbah2016deep}. Notably, Pollack \textit{et al.} \cite{pollack2022axon} incorporates topological information of axon into the loss function to train the deep learning model in order to provide better voxel-based segmentation and axon centerline detection. Motta \textit{et al.} \cite{motta2019dense} takes a semi-automated approach to conduct the axon segmentation task in the mouse cortex. And in \cite{plebani2022high}, a high-throughput deep learning pipeline is proposed to tackle the more challenging problem of unmyelinated axon instance segmentation. There also exists a public dataset called AxonEM \cite{wei2021axonem}, which is a large-scale EM dataset from the human and mouse cortex that provides dense ground truth instance segmentation of axons.

However, almost all previous work on axon segmentation (such as those mentioned above) has been based on EM images. Thus, currently there are no publicly available XNH datasets of white-matter axons. Compared to previous EM segmentation benchmarks (AxonEM \cite{wei2021axonem} and CREMI\cite{cremi2miccai}, which focus on gray matter (neuropil), XNH white-matter data presents unique challenges. XNH data has lower resolution and contrast (tradeoffs for much higher imaging rates), which makes membrane detection more difficult. Myelinated axons are densely packed into tracts but generally have a linear geometry. As a result, each axon can be efficiently and quickly annotated via skeleton tracing, rather than labeling every voxel. 

To stimulate new research to address these challenges, we have created a benchmark for 3D segmentation of white-matter axons from XNH data: X-ray Projectomic Reconstruction: Extracting Segmentation from Skeletons (XPRESS). This benchmark consists of three (training, validation, test) XNH volumes from the white matter region in the mouse brain, each containing manual annotations for myelinated axon trajectories (Fig. \ref{fig:dataset}b,c).  Additionally, the training set contains a sub-volume with voxel-wise manual annotations (segmentation). The goal of the challenge is to use the annotations on the training/validation datasets (voxel-wise and/or skeletons) to train a 3D instance segmentation model. The results will be evaluated on the test dataset by comparing the consistency between the axon segmentation and ground truth skeleton annotations.

\textbf{Contributions}
Our contributions are as follows, 
First, we provide the first publicly available, annotated XNH dataset of white-matter axons in the mouse brain to enable development of scalable reconstruction algorithms for long-range connectomics (projectomics). Second, we introduce a novel task to utilize efficient skeleton-based annotations for 3D instance segmentation. Third, we present and evaluate a baseline model for this task which demonstrates feasibility and highlights the potential for improvements.

%% file: 3-data-challenge.tex
\section{Data and Challenge}
\label{sec:dataset}

\subsection{Dataset Description}

The three datasets (training, validation, and test) are each $1200^3$ voxels with $33.3$ nm isotropic voxel size, corresponding to a physical volume of $(40 \mu m)^3$. The data were imaged with similar techniques and parameters as previously published data in cortical neuropil \cite{kuan2020}, but was collected from cortical white matter, which consists of densely packed myelinated axons. The images were upsampled from raw data collected with 100 nm voxel size to make the voxel size more similar to EM datasets (typically 4-20 nm voxels).


\subsection{Dataset Annotation}

For each volume, we manually traced axons using the WebKnossos interface \cite{boergens2017webknossos} to create a skeleton annotation (directed graph) for each axon. For the training and test datasets, subvolumes of $1100^3$ voxels were annotated, while in the validation set a subvolume of $700^3$ voxels was annotated. These annotations include a total of 1815 individual axons. Additionally, we provide voxelwise segmentation annotations for a $200^3$ voxel region of the training volume. White-matter axons are generally large-caliber ($>1$ um) and high-contrast (due to the myelin sheaths), making it possible to manually annotate with high accuracy. However, due to the moderate XNH resolution (recorded with 100 nm voxels, corresponding to ~200 nm true resolution \cite{kuan2020}), some locations may be ambiguous. To produce high quality ground truth, all annotations were reviewed by a second annotator, and we conservatively left out axons which contained difficult or ambiguous continuations.

\subsection{Challenge Description}

For the XPRESS challenge, the goal is to segment the test dataset such that voxels corresponding to each myelinated axon are labeled by the same segment ID, and voxels corresponding to different axons are labeled by different segment IDs. The training and validation datasets and ground truth annotations are provided as training data. The submission format is an image volume in which the value of each voxel is a segment ID. These segmentations will be compared with the ground truth tracing to calculate the accuracy score. 

Since most segmentation algorithms currently require voxel-wise ground truth (rather than skeletons) for initial training, we have provided a limited amount of voxelwise GT. We envision that participants will begin training on the voxel-wise ground truth, and then augment the training using the larger volume of skeleton GT. However, it is also possible to train on skeleton or voxel-wise GT only. The participants will have flexibility to use either or both of the provided annotations to train their models, and submit the voxel-wise prediction on the test volume.

%% file: 4-baseline-evaluation.tex
\section{Baselines and Evaluation}
\begin{figure*}[t]
\begin{center}

\includegraphics[width=0.70\linewidth]{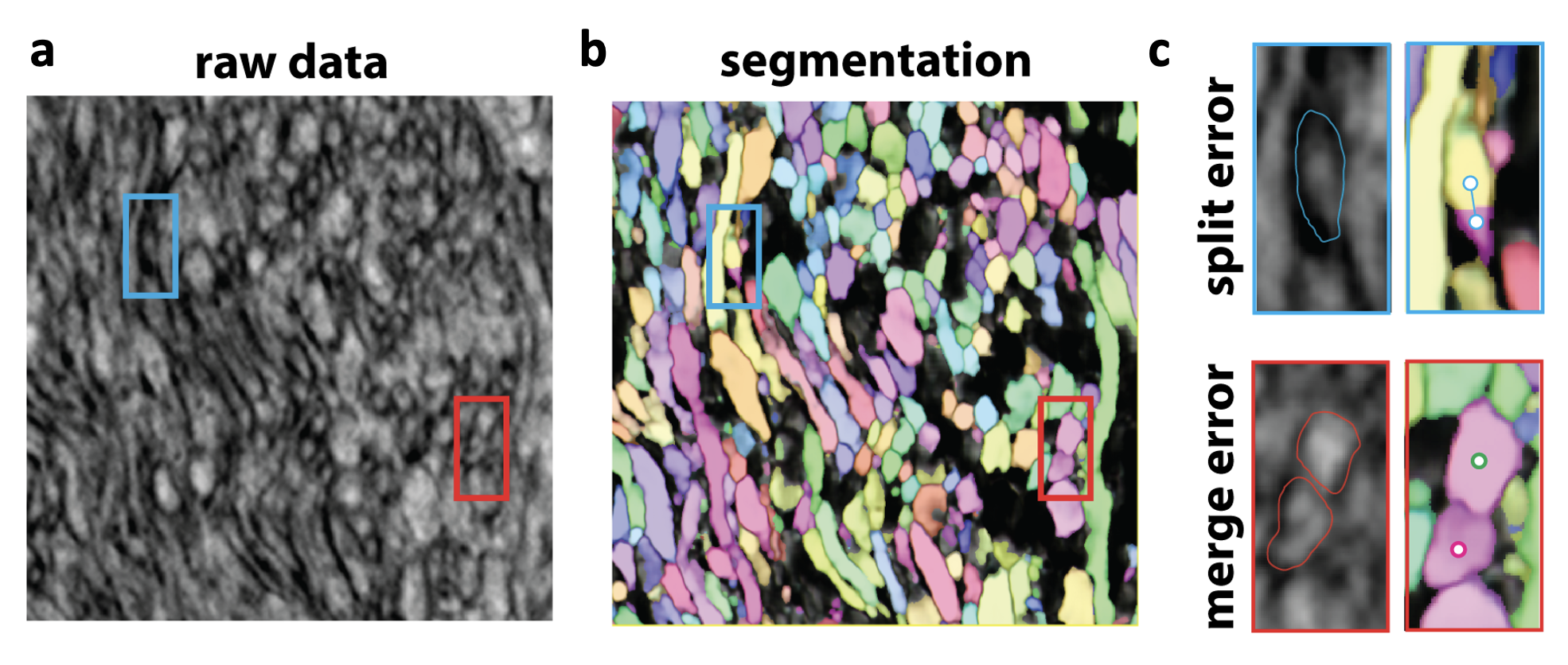}
\end{center}
\vspace{-10pt}
\caption{\textbf{(a)} XNH image data corresponding to the same region show in (b). Blue and rectangles correspond to regions shown in (c). \textbf{(b)} Example axon segmentation from the \texttt{UNET-vxl} model. Different segments are shown in different colors, overlaid over the predicted affinity map, which can be interpreted as a boundary prediction. \textbf{(c)} Example of a split error and merge error. Left - XNH data. Right - Segmentation prediction showing one axon (GT) split into two different predicted segments, and two
different axons (GT) merged into the same predicted segment. Images correspond to blue box and red box in (a). }
\label{fig:eval}
\end{figure*}


\subsection{Baseline Model}

To establish a baseline, we trained U-Net model (Fig. \ref{fig:eval}, Table \ref{table:erl}) previously developed for segmentation of EM data~\cite{funke2018large,nguyen2022structured}, further refined for X-ray segmentation~\cite{kuan2020} and augmented with long-range affinties~\cite{lee2017superhuman}. This model was trained only on the voxel-wise ground truth subset of the training set. The model consists of 3 layers, with each layer performing a [2,2,2] downsampling factor and a doubling of feature maps starting with 24 features. The output was further convolved and passed through a sigmoid activation to get 12 long-range affinities (x-1,y-1,z-1,x-3,y-3,z-3,x-9,y-9,z-9,x-27,y-27,z-27, but only x-1,y-1,z-1 are used for segmentation) per voxel. The network used MSE loss and optimized with an Adam optimizer with a learning rate of 2.5e-5. The predicted affinties are then post-processed into segmentation using Segway~\cite{kuan2020,nguyen2022structured}, a segmentation pipeline based on Daisy~\cite{daisy2022github}.

\subsection{Training}

We used a training pipeline based on Gunpowder~\cite{gunpowder} for training. On each training batch, Gunpowder randomly picks a 132,132,132 patch from the training volumes and performs data augmentation consisting of random rotation, mirroring, transposing, intensity, noise, and elastic transformations. We initially trained the model to 1,000,000 iterations and then optimized hyper-parameters using validation as described in the following section.

\subsection{Evaluation Metric}


Automated segmentation can contain two types of errors: (1) split errors occur if two nodes from the same skeleton intersect segments with different segment IDs (Fig. \ref{fig:eval}d), and (2) merge errors occur if two nodes from different skeletons intersect the same segment (Fig. \ref{fig:eval}e). Generally, there is a trade-off between split and merge errors. In the baseline model, the agglomeration threshold hyper-parameter, which determines how aggressively different supervoxels are combined together into segments, parametrizes this trade-off.

To evaluate split/merge performance as a single score, we used a combination of expected run-length (ERL)~\cite{januszewski2018high} and split and merge Rand indices~\cite{rand1971objective}, which are commonly used metrics for segmentation evaluation~\cite{sheridan_local_2022}. The following formula computes the combined XPRESS score:

\begin{equation}
norm\_erl = \frac{ERL}{skeleton\_length}
\end{equation}

\begin{equation}
    rand = \frac{rand\_split + rand\_merge}{2}
\end{equation}

\begin{equation}
    xpress\_score = \frac{norm\_erl + rand}{2}
\end{equation}

The rationale for combining multiple metrics is that each metric penalizes certain types of errors in different ways. ERL measures the average length of expected error-free segments, but heavily penalizes merge errors by counting any segments containing merge errors as 0 length. This reflects the view that  proofreading merge errors is much harder than split errors~\cite{januszewski2018high}. However, recent approaches to proofreading where shortest path-length between merge objects can be efficiently calculated~\cite{dorkenwald2022flywire} or where segmentation is chunked into small blocks to avoid catastrophic merge errors~\cite{nguyen2022structured} can reduce the burden of correcting merge errors. Thus, the split and merge clustering metrics (Rand index) are included in XPRESS score as well.



\begin{table}[t]
\centering  
\renewcommand{\arraystretch}{1.5}
\caption{Evaluation results for baseline model}
\begin{tabular}{c|ccc|c}
\hline
         & Normalized ERL & Rand split & Rand merge & XPRESS \\  \hline
        Validation & 0.6667 & 0.7122 & 0.9543 & 0.7500  \\ \hline
        Test & 0.7428 & 0.7276 & 0.9572 & 0.7926\\
         \hline
       
\end{tabular}
\label{table:erl}
\end{table}

\subsection{Analysis}

The baseline model produces reasonable membrane predictions (affinities) (Fig. \ref{fig:eval}b,c) that are much cleaner than the noisy input x-ray images. It is far from perfect, however, with a significant amount of split errors likely a result of insufficient training data. We expect that utilizing skeleton ground-truth in training would further increase accuracy.

Furthermore, through this challenge, we anticipate that innovations in network architecture, loss functions, agglomeration algorithms, or other aspects of the segmentation pipeline will boost performance beyond the baseline model presented here. To this end, we are providing checkpoints of the pre-trained baseline model, its output affinities, as well as its output segmentation for participants to work on without having have to train a new model from scratch.

%% file: 5-plan-schedule.tex
\section{Plan \& Schedule}
\label{sec:plan&schedule}
\bfsection{Challenge schedule} The schedule for the XPRESS challenge is shown in Table~\ref{table: schedule}. Download links, challenge details and submission instructions are available on the challenge website: https://xpress.grand-challenge.org.   We will also provide google colab and neuroglancer links to help visualize the datasets and baseline results. The participants will submit the voxel-wise prediction on the test dataset from their methods to the website. The submitted results will be evaluated based on our provided metrics (XPRESS score). The source code for the method is encouraged but not required. We will give participants the option to put the link to their code on the leaderboard page. All results will be announced publicly through a public leaderboard. 

\bfsection{Challenge award} After the challenge deadline, a certificate will be awarded to challenge top-3 teams (1 winner and 2 runners-up). They are required to submit a 4-page solution. At the workshop, all participants will have an opportunity to give poster presentations and the top 3 teams will be invited for oral presentations. The top 3 or more teams will also be invited for writing an overview paper. Additionally, the participating teams may publish their own results separately. 

\bfsection{Post-challenge plan} After the ISBI conference, the challenge will continue to be open. We will not release the test case labels to the public. Only the organizers have access to the test case labels, during and after the challenge.

\begin{table}[t]
\centering
\small
\caption{XPRESS challenge schedule.}
\begin{tabular}{l|l}
\hline
Date & Schedule \\ \hline
Feb 6, 2023 & Registration starts, release of datasets\\
Feb 6, 2023 & Submission of results accepted \\
Mar 30, 2023 & Deadline for submissions \\
Apr 1, 2023 & Final leaderboard announced \\ 
Apr 18, 2023 & On-site workshop presentation at ISBI 2023 \\

\hline
\end{tabular}

\label{table: schedule}
\end{table}

%% file: 7-reference.tex
\bibliographystyle{IEEEtran}
\bibliography{refs.bib}

%% file: 6-biosketch.tex
\section{Acknowledgements}
\label{sec:acknowledgemetns}

This project received funding from the NIH (EB032217) to ATK and (MH128949) to WCAL and AP, and the European Research Council (ERC) under the European Union's Horizon 2020 research and innovation programme (grant agreement n°852455) to AP. We acknowledge the ESRF for granting beamtime for proposal LS2892. 

\section{Biosketch of Organizers}
\label{sec:biosketch}

Tri Nguyen is a Postdoctoral Fellow at Harvard Medical School, working with Wei-Chung Lee on developing novel techniques for automated segmentation and analyzing a connectome of the mouse cerebellum. Previously he obtained a Ph.D. at Princeton University in computer architecture.

Mukul Narwani is a recent graduate from Northeastern University with a background in physics, computer science and math. He is currently working as a data scientist in the Lee Lab at Harvard Medical School. He has developed tools for CT image data augmentation and has experience on EM neuron annotation for both recreating cortical neuron connectivity and generating cerebellum ground truth.

Mark Larson is an undergraduate student at Northeastern University studying math and physics, working as a co-op student in the Lee Lab at Harvard Medical School. Before working on the XPRESS challenge, he had more than 6 months of neuron annotation experience reconstructing cortical neurons from electron microscopy data. 

Yicong Li is a Ph.D. student in computer science at Harvard University, advised by Hanspeter Pfister. He is working on biomedical image analysis, AI for healthcare,  computer vision, and connectomics. He has research papers published in top venues including CVPR, MICCAI, etc. Previously, he obtained his master's degree at Tsinghua University and bachelor's degree at Sichuan University.

Hanspeter Pfister is An Wang Professor of Computer Science in the School of Engineering and Applied Sciences at Harvard University. Before joining Harvard he worked for over a decade at Mitsubishi Electric Research Laboratories where he was Associate Director and a Senior Research Scientist. Dr. Pfister has a Ph.D. in Computer Science from the State University of New York at Stony Brook and an M.S. in Electrical Engineering from ETH Zurich, Switzerland. He was also a co-organizer for ISBI 2021 MitoEM challenge.

Donglai Wei is an Assistant Professor in the Computer Science Department at Boston College. His research focuses on developing novel registration and reconstruction algorithms for large-scale (currently petabyte-scale) connectomics datasets to empower neuroscience discoveries. During his Ph.D. at MIT under Prof. William Freeman, he worked on video understanding problems, including arrow of time and Vimeo-90K benchmark. Since his postdoc at Harvard University, he has embarked on the quest to reconstruct the brain's wiring diagram. He was the lead organizer for ISBI 2021 MitoEM challenge and co-organizer for MICCAI 2020 RibFrac Challenge.

Nir Shavit worked on multicore algorithms before multicore machines existed. In the past 8 years, he has shifted his interests to connectiomcs, and the role of parallelism in neural networks, both real and artificial. He is a fellow of the ACM, a winner of the 2004 Gödel Prize in theoretical computer science, the 2012 Dijkstra Prize in Distributed Computing, and is the author of The Art of Multiprocessor Programming. Nir leads the Computational Connectomics group at MIT-CSAIL.

Lu Mi is a Ph.D. student at MIT CSAIL, working under the supervision of Nir Shavit. Her research involves developing fast automatic connectomics pipeline to discover the brain, and building models to bridge the gap between the anatomical structure and function of the brain. She is an incoming Shanahan Foundation Fellow at the Allen Institute and postdoctoral researcher at the University of Washington, Seattle.

Alexandra Pacureanu is a Scientist at the European Synchrotron leading the X-ray Neuroimaging at Nanoscale unit. Her research path started at the Biomedical Imaging Laboratory (Creatis) in Lyon, followed by the Centre for Image Analysis and Science for Life Laboratory in Uppsala. She then continued her work at the interface between X-ray microscopy, image analysis and biomedical applications at the ESRF, Harvard Medical School and University College London, before establishing her research group in France. She has been part of the ISBI community since 2009. 

Wei-Chung Allen Lee is an Associate Professor of Neurology at Boston Children’s Hospital and Harvard Medical School. He has over 10 years of experience in “functional connectomics”, applying imaging approaches to understand computation and information processing in neuronal circuits. He was awarded the Society for Neuroscience’s inaugural Jennifer N. Bourne Prize, for advances in understanding brain structure-function relationships at the nanometer scale.

Aaron Kuan is a NIH NIBIB K99 Postdoctoral Fellow at Harvard Medical School, working under the supervision of co-mentors Wei-Chung Lee and Christopher Harvey. His research involves developing connectomic imaging and analysis technologies for studying cognition in the mammalian cortex. In recent work, he demonstrated that XNH can be applied to connectomics, providing a method to obtain brain-wide projectomic maps.